\documentclass[runningheads]{llncs}
\usepackage{graphicx}
\usepackage[hidelinks]{hyperref}

\usepackage{wrapfig}
\usepackage[strings]{underscore}

\begin{document}
\title{Investigating Conversational Search Behavior For Domain Exploration}
%
%
\author{Phillip Schneider\inst{1} \and
Anum Afzal\inst{1} \and
Juraj Vladika\inst{1} \and \\
Daniel Braun\inst{2} \and
Florian Matthes\inst{1}
}

\authorrunning{P. Schneider et al.}
%
\institute{Department of Computer Science, Technical University of Munich, Boltzmannstrasse 3, Garching, Germany\\
\email{\{phillip.schneider, anum.afzal, juraj.vladika, matthes\}@tum.de} \and
Department of High-tech Business and Entrepreneurship, University of Twente, Drienerlolaan 5, Enschede, The Netherlands\\
\email{d.braun@utwente.nl}}

\maketitle              
\begin{abstract}
Conversational search has evolved as a new information retrieval paradigm, marking a shift from traditional search systems towards interactive dialogues with intelligent search agents. This change especially affects exploratory information-seeking contexts, where conversational search systems can guide the discovery of unfamiliar domains. In these scenarios, users find it often difficult to express their information goals due to insufficient background knowledge. Conversational interfaces can provide assistance by eliciting information needs and narrowing down the search space. However, due to the complexity of information-seeking behavior, the design of conversational interfaces for retrieving information remains a great challenge. Although prior work has employed user studies to empirically ground the system design, most existing studies are limited to well-defined search tasks or known domains, thus being less exploratory in nature. Therefore, we conducted a laboratory study to investigate open-ended search behavior for navigation through unknown information landscapes. The study comprised of 26 participants who were restricted in their search to a text chat interface. Based on the collected dialogue transcripts, we applied statistical analyses and process mining techniques to uncover general information-seeking patterns across five different domains. We not only identify core dialogue acts and their interrelations that enable users to discover domain knowledge, but also derive design suggestions for conversational search systems.

\end{abstract}

\section{Introduction}
Driven by major advances in natural language processing, the ubiquitous availability of conversational agents ushered in a new era for human-computer interfaces. In consequence, interactions between humans and machines shift towards the medium of language \cite{klopfenstein2017rise,mctear2016conversational}. Even though modern conversational agents have a broad skill set in following task-oriented commands or engaging in short chit-chat conversations, their information-seeking capabilities are predominantly confined to answering factoid questions. A limitation of the question-answering paradigm is its inherent dependence on the users' prior knowledge, which is the prerequisite for being able to ask meaningful questions in the first place \cite{belkin1980anomalous,furnas1987vocabulary}. Especially in exploratory search scenarios, where users with unclear information goals are confronted with unfamiliar domains, it is necessary to support search behaviors that go beyond simple query-response interactions \cite{white2009exploratory}. Hence, there is a growing research interest in multi-turn conversational search systems. While some scholars approach this topic by developing theories and conceptual frameworks \cite{azzopardi2018conceptualizing,radlinski2017theoretical,thomas2021theories}, others perform laboratory studies in combination with dialogue analysis to ground models of search behavior in empirical observations \cite{trippas2020towards,vakulenko2020conversational,vtyurina2017exploring}.

However, most existing laboratory studies focus only on experimental setups with search scenarios that are not exploratory in nature but constrained by predefined information needs and search tasks. Therefore, we designed a study for answering the research question: \textit{What is the characteristic dialogue structure of information-seeking conversations for domain exploration?} As far as we know, our experiment is the first to collect transcripts of completely open-ended exploratory search dialogues in five domains. Our contributions are twofold: (i) We publish an annotated corpus of exploratory search dialogues with five domain datasets.\footnote{Repository: \href{https://github.com/sebischair/conversational-domain-exploration-data}{https://github.com/sebischair/conversational-domain-exploration-data}} (ii) We identify core dialogue acts and domain-independent dialogue flow patterns which can inform the design of conversational systems.

\section{Related Work}
In the literature on conversational systems, dialogue analysis is common research practice. It facilitates the conceptual understanding of human behavior by means of examining communication patterns, information flows, or vocabulary choice \cite{bunt1999dynamic,yankelovich2008using}. Concerning information search, dialogue analysis has been applied to characterize information-seeking conversations and develop theoretical models \cite{vakulenko2021large}. While some researchers gather dialogue data from natural settings, such as reference interviews or online support platforms \cite{daniels1985using,qu2018analyzing,saracevic1997users}, others conduct controlled laboratory studies to set up an artificial search scenario, as is the case in our experiment. Vtyurina et al. \cite{vtyurina2017exploring} performed an experiment to explore users' preferences in solving search tasks with three kinds of assistants: a chatbot, a human, and a perceived automatic system simulated by a human. A study carried out by Trippas et al. \cite{trippas2018informing} investigated how users communicate in an audio-only search setting. Both experiments had clearly defined search tasks and information needs assigned to the participants, which were not exploratory but goal-oriented. Another related work is from Vakulenko et al. \cite{vakulenko2020conversational}, where students engaged in conversations for exploratory browsing to find a specific dataset in an open data portal. In contrast, we do not restrict the search to a predefined task, but instead, we only instructed participants to explore an unknown dataset. A further distinction is that we record multiple dialogue sessions across five domains and compare general interaction patterns of exploratory search.

\section{Method}
A total of 26 participants took part in the study, which was conducted in English. The participants were university students recruited from a teaching course. All students had previous experience with chat interfaces, but no prior knowledge of the datasets they were instructed to explore. We scraped five publicly available datasets from the internet. All of them have a relational structure, in which a set of items is characterized by a set of attributes. The tabular datasets were selected by two aspects. For one thing, they had to be licensed under Creative Commons BY-SA 3.0 or BY-SA 4.0, and for another, they had to contain enough interesting data items for an engaging conversation. Ultimately, we acquired one dataset for each of the following domains: geography, history, media, nutrition, and sports. Table~\ref{tab:domain-datasets} in the Appendix lists each of the five datasets along with a short description of their content. 

The experimental setup consisted of 26 chat sessions between two participants, where one participant acted as an information seeker and the other as an information provider. Based on personal preference, every student was given one dataset in the form of a spreadsheet as an information source for the provider role. We grouped the students into pairs with two distinct datasets, ensuring mutual interest in each other's domain. Each pair was assigned to a text-based chatroom. Seekers were only instructed to explore and inquire information about the unknown dataset of their partner, but no concrete search task was specified. After one session of 15 minutes, students in the seeker role completed a feedback survey. It contained two free-form questions about unmet information needs as well as suggestions to improve the search experience. After completing the feedback survey, the participants switched roles and started with their second chat session regarding the other domain.

After running the experiment, the dialogue scripts were annotated. This task of dialogue act annotation identifies the function or goal of a given utterance \cite{sinclair1975towards}. Two researchers independently labeled each message with a speech act and corresponding dialogue act. To assess the reliability of the inter-annotator agreement, we calculated Cohen’s Kappa coefficient \cite{cohen1960coefficient}. The annotations of speech and dialogue acts had coefficients of 0.93 and 0.86, respectively. As suggested by Cohen, Kappa values from 0.81 to 1.00 indicate almost perfect agreement. To come up with a suitable group of dialogue acts, we started with an initial set derived from the widely known taxonomy of Bach and Harnish \cite{bach1979linguistic}. Through regular discussions, we clarified ambiguous labels and resolved disagreements between the annotators. Thereby, the set of used dialogue acts evolved through adding or removing certain acts to better fit the dialogue corpus.

For examining the annotated corpus, we calculated various descriptive statistics. Furthermore, we employed process mining techniques since they have been successfully applied to discover sequential patterns from event logs in various data formats, including conversational transcripts. We chose a Python-based state-of-the-art process mining library called PM4Py.\footnote{PM4Py process mining package: \href{https://pm4py.fit.fraunhofer.de}{https://pm4py.fit.fraunhofer.de}} 

\section{Results and Discussion}
\subsubsection{Statistical Analysis.}
We performed a statistical analysis to describe the occurrence of linguistic constituents like speech or dialogue acts. The annotated dialogue corpus contains 669 individual messages from 26 sessions. Table~\ref{tab:summary-statistics} lists the most important summary statistics. Looking at the different domains, we see that the minimum and maximum message count varies greatly, ranging from a chat with 8 messages to a more extensive chat with 62 messages. Participants exchanged on average 25.7 text messages with each other, which is significantly higher than in the more goal-directed conversational search experiment from Vakulenko et al. \cite{vakulenko2020conversational}. Considering the verbosity of the messages, the mean length of the utterances is 46.4 characters. The standard deviation of text length was unusually large for history, due to a single very long message. Besides this outlier, no irregularities in the dataset were found. It can be noted that history and sports are the domains with not only the lowest message count and the smallest average of messages per session, but they also have the shortest messages on average when excluding the outlier in the history domain.

\begin{table}
\centering
  \caption{Summary statistics of dialogue corpus.}
  \label{tab:summary-statistics}
  \resizebox{\textwidth}{!}{ 
  \begin{tabular}{|l|c|c|c|c|c|c|}
    \hline
    Aspect & Geography & History & Media & Nutrition & Sports & Overall\\
    \hline
    \scriptsize Number of messages & \scriptsize 169 & \scriptsize 98 & \scriptsize 156 & \scriptsize 153 & \scriptsize 93 & \scriptsize 669\\
    \scriptsize Number of sessions & \scriptsize 6 & \scriptsize 5 & \scriptsize 5 & \scriptsize 5 & \scriptsize 5 & \scriptsize 26\\
    \scriptsize Min-max messages per session & \scriptsize 11-41 & \scriptsize 8-29 & \scriptsize 20-62 & \scriptsize 21-45 & \scriptsize 16-22 & \scriptsize 8-62\\
    \scriptsize Average messages per session & \scriptsize 28.2 & \scriptsize 19.6 & \scriptsize 31.2 & \scriptsize 30.6 & \scriptsize 18.6 & \scriptsize 25.7\\
    \scriptsize Average characters per message & \scriptsize 47.3 & \scriptsize 48.5 & \scriptsize 45.0 & \scriptsize 49.0 & \scriptsize 40.9 & \scriptsize 46.4\\
  \hline
\end{tabular}
}
\end{table}

\begin{wrapfigure}{r}{0.5\textwidth}
  \center
  \includegraphics[clip, trim=0.2cm 0.2cm 0.2cm 0.2cm, width=0.5\textwidth]{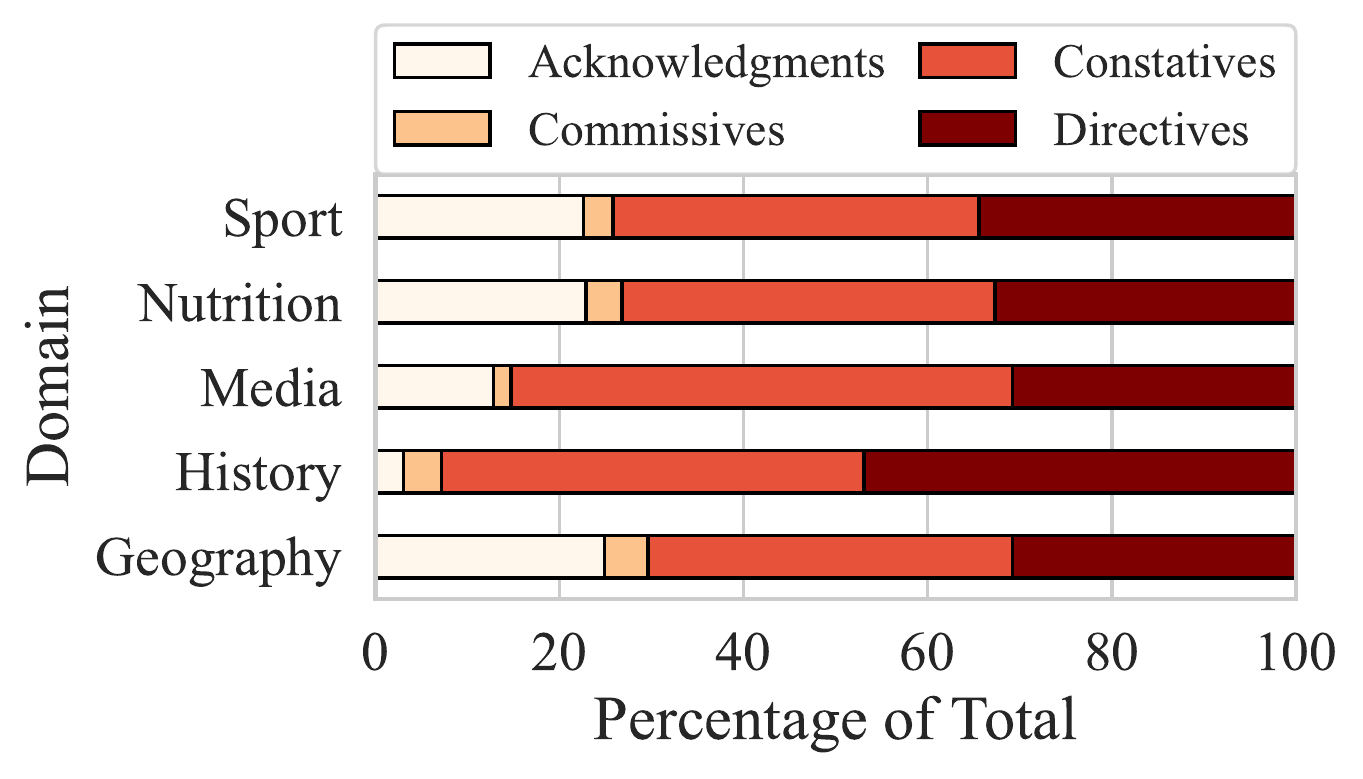}
  \caption {Distribution of speech acts.}
  \label{fig:speech-act-stack}
\end{wrapfigure}

More insights about the linguistic elements of the dialogue transcripts can be gained by comparing the distribution of speech acts. As proposed by Bach and Harnish \cite{bach1979linguistic}, there are four groups of illocutionary speech acts: Constatives express an intention to convey information. Directives express the intention to get the addressee to do something. Commissives are acts of obligating oneself to do something. Acknowledgments express mutual understanding or attitudes that are expected on particular occasions. Figure~\ref{fig:speech-act-stack} illustrates the distribution of observed speech acts across the five domains. In total, constatives (44.2\%) and directives (34.1\%) are most predominant, accounting for over three-quarters of all speech acts. Acknowledgments and commissives make up 18.1\% and 3.6\%, respectively. Considering all domains, constatives have a higher occurrence than directives, followed by acknowledgments and commissives. The history sessions deviate from this rule since the ratios of constatives and directives are almost equal and there are only very few acknowledgments.

\begin{table}
  \caption{Distribution of dialogue acts sorted by relative frequency.}
  \label{tab:dialogue-acts}
  \resizebox{\textwidth}{!}{ 
  \begin{tabular}{|l|l|c|l|l|}
    \hline
    Speech Act & Dialogue Act & Percentage & Definition\\
    \hline
    \scriptsize Directives & \scriptsize Request & \scriptsize 32.0\% & \scriptsize Express a general information need.\\
    \scriptsize Constatives & \scriptsize Describe & \scriptsize 19.3\% & \scriptsize Provide a description of an information item.\\
    \scriptsize Acknowledgments & \scriptsize Acknowledge & \scriptsize 8.1\% & \scriptsize Express that an utterance was understood.\\
    \scriptsize Constatives & \scriptsize List & \scriptsize 7.0\% & \scriptsize List multiple information items from the data.\\
    \scriptsize Constatives & \scriptsize Rank & \scriptsize 6.7\% & \scriptsize Rank information items by a given metric.\\
    \scriptsize Acknowledgments & \scriptsize Greet & \scriptsize 5.8\% & \scriptsize Open the conversation with an initial greeting.\\
    \scriptsize Constatives & \scriptsize Count & \scriptsize 4.6\% & \scriptsize Count the number of a specified set of items.\\
    \scriptsize Acknowledgments & \scriptsize Thank & \scriptsize 4.2\% & \scriptsize Express gratitude with regard to a response.\\
    \scriptsize Constatives & \scriptsize Verify & \scriptsize 3.7\% & \scriptsize Verify if the dataset contains a specific item.\\
    \scriptsize Commissives & \scriptsize Offer & \scriptsize 3.0\% & \scriptsize Offer options for information exploration.\\
    \scriptsize Constatives & \scriptsize Accept & \scriptsize 2.4\% & \scriptsize Agree with a suggested exploration direction.\\
    \scriptsize Directives & \scriptsize Clarify & \scriptsize 2.1\% & \scriptsize Ask a clarifying question for a given utterance.\\
    \scriptsize Commissives & \scriptsize Promise & \scriptsize 0.6\% & \scriptsize Promise to perform a requested action.\\
    \scriptsize Constatives & \scriptsize Reject & \scriptsize 0.4\% & \scriptsize Disagree with a suggested exploration direction.\\
  \hline
\end{tabular}
}
\end{table}

Dialogue acts are specialized speech acts that depend on the conversation setting. Thus, they allow for more granular linguistic analysis. We present an overview of all identified dialogue acts in Table~\ref{tab:dialogue-acts}. Expressing an information request is the most frequent act, which accounts for every third utterance, whereas describing a data item ranks second with 19.3\%. Together, these two dialogue acts represent already half of all interactions. Other information-providing dialogue acts, such as listing, ranking, verifying, or counting, claim a significant share as well. It is also noteworthy that after a suggested exploration direction, seekers acted receptive, accepting over five times more often than rejecting. With regard to the speech act of acknowledgments, Table~\ref{tab:dialogue-acts} shows that acknowledging and thanking are important functions for communicating feedback. As can be seen from the relatively few offer (3.0\%) and promise (0.6\%) dialogue acts, the participants used commissives rarely. The same holds true for asking clarification questions. Overall, the vast majority of utterances serve the functions of requesting information, providing information, and giving feedback.

\subsubsection{Process Mining Analysis.}
\begin{figure}
  \center
  \includegraphics[width=\textwidth]{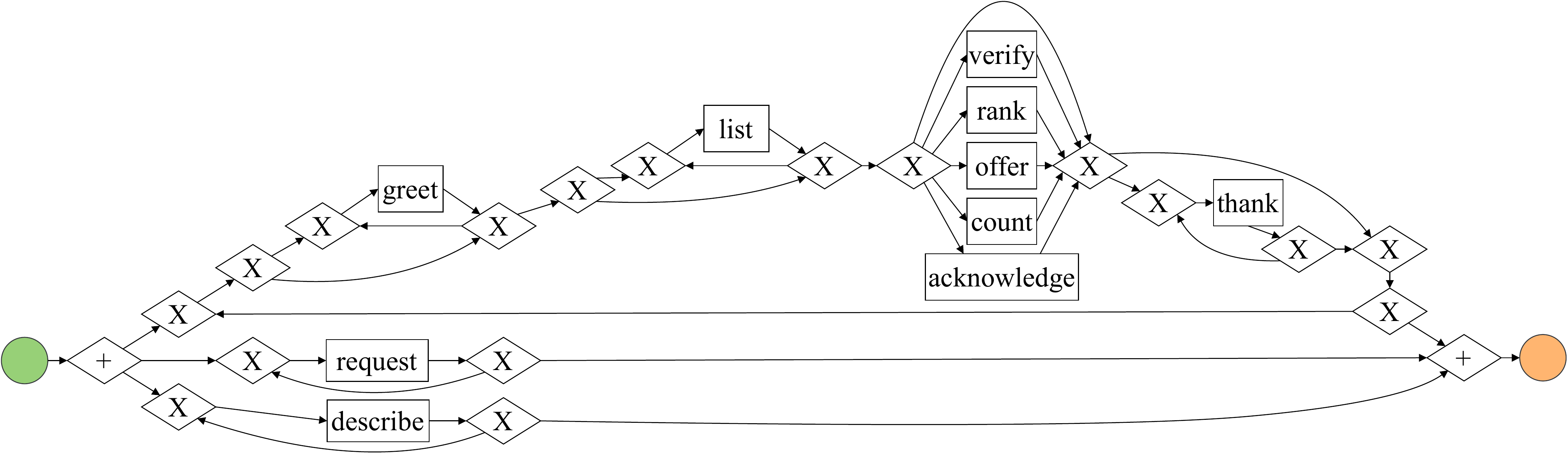}
  \caption{Process model of extracted dialogue sequence flow.}
  \label{fig:dialogue-act-bpmn}
\end{figure}

Aside from identifying dialogue acts, we applied an inductive miner algorithm to discover the underlying core process for exploratory search. Figure~\ref{fig:dialogue-act-bpmn} depicts the extracted domain-independent dialogue sequence flow which manifested itself in all five domains. The nodes correspond to the ten most frequent dialogue acts. The arrows show the direction of the conversation flow. Lastly, there are diamond-shaped gateway symbols, an exclusive gateway (X) breaks the flow into mutually exclusive flows, and an inclusive gateway (+) represents concurrent flows. Concurrency means that the order of dialogue acts varied between the analyzed conversation transcripts. From Figure~\ref{fig:dialogue-act-bpmn}, we can discern that the dialogue logs consist of three concurrent main loops. One is for requesting information, a second is for describing information items, and a third incorporates a sequence of the remaining dialogue acts. The two former loops are indispensable since they are used at least once in every conversation, whereas the third loop allows for skipping specific dialogue acts. 

Considering the dialogue transcripts across all domains, we observed that most chat sessions started with a greeting and an ensuing question from the information seeker. This request usually came in the form of asking what the dataset is about in general, followed by asking about its different dimensions. The information provider tried to fulfill these requests by listing column names or exemplary data records. On that basis, the seeker proceeded with asking more detailed questions, often about aggregated, or sorted data items, which relate to count and rank dialogue acts. In the related study from Vakulenko et al. \cite{vakulenko2020conversational}, similar dialogue acts for browsing through relational data structures were identified. These acts help information seekers to better understand the scope of the dataset by getting to know the existing attributes along with their facets. When the provider had trouble finding an answer, the provider sometimes offered alternative options. The seeker, in turn, gave feedback by acknowledging utterances or expressing gratitude for relevant responses. The participants commonly ended their chats by thanking each other.

\subsubsection{Discussion of Design Suggestions.}
Our analysis demonstrated the key role of certain dialogue acts in domain exploration scenarios. Effective search agents must be able to emulate these communication functions and handle the dialogue flow outlined hereinafter. Concerning the interplay between constatives and directives, we found that the observed dialogues are seeker-driven with every third dialogue act being a request. Information providers addressed these requests by first listing or describing metadata, along with counting, ranking, and verifying information items. This iterative interaction process supported seekers in building a mental model of the dataset’s dimensionality, which was essential for asking more complex follow-up questions. This is in line with Belkin's Anomalous State of Knowledge model (ASK) \cite{belkin1980anomalous} because it is only possible for seekers to formulate an information need if they have first perceived a knowledge gap. Providers were guided in their informative responses through the seekers’ acknowledgments.

A similar picture emerged from the feedback surveys of the participants. Overall, they had a positive impression of their chat sessions, although when asked about suggestions for improvement, they pointed towards both the chat interface and the dialogue interaction. For instance, they wanted visual data summaries in the chat window, clickable buttons with predefined replies, or timestamped messages. Regarding the interaction, they demanded a short introduction of the dataset, i.e., metadata, right at the start of the session. This prior knowledge was an essential requirement for engaging in a deeper exploration of the domain. Also, they criticized one-word answers, wishing for more descriptive responses and question suggestions from the provider. Some of these user preferences were also observed by Vtyurina et al. \cite{vtyurina2017exploring}. In consequence, it seems evident that exploratory conversational search depends on a proactive search agent, which can quickly familiarize the user with the explored domain.

\section{Conclusion and Future Work}
In this paper, we presented a laboratory study designed to investigate conversational search behavior in the context of discovering unfamiliar domains. Our empirical analysis not only revealed core dialogue acts that information seekers use, but also a domain-independent dialogue act flow sequence. In addition, we believe that our derived design suggestions are vital for a user-centered design of conversational agents. Two major limitations of our study arise due to having a biased participant sample of only university students and focusing exclusively on tabular datasets. In future work, we plan to use our insights to build and evaluate exploratory search agents for more diverse user groups with the capability to retrieve information from different data sources and structures.

\appendix
\section{Appendix}
\begin{table}[h]
\centering
  \caption{Overview of the five domain datasets used in the experiment.}
  \label{tab:domain-datasets}
  \begin{tabular}{|l|c|c|l|}
    \hline
    Domain & \# Rows & \# Columns & Short Description\\
    \hline
    \scriptsize Geography & \scriptsize 98 & \scriptsize 5 & \scriptsize Geographic information about nature parks.\\
    \scriptsize History & \scriptsize 11341 & \scriptsize 17 & \scriptsize Biographic data about historical figures.\\
    \scriptsize Media & \scriptsize 500 & \scriptsize 5 & \scriptsize Data about time travel literature, films, and TV series.\\
    \scriptsize Nutrition & \scriptsize 285 & \scriptsize 9 & \scriptsize Nutritional values of common food products.\\
    \scriptsize Sports & \scriptsize 77 & \scriptsize 6 & \scriptsize Data about international football records.\\
  \hline
\end{tabular}
\end{table}

\subsubsection{Acknowledgments.}
This work has been supported by the German Federal Ministry of Education and Research (BMBF) Software Campus grant 01IS17049.

%
%
%
\bibliographystyle{splncs04}
\bibliography{mybibliography}

\begin{thebibliography}{10}
\providecommand{\url}[1]{\texttt{#1}}
\providecommand{\urlprefix}{URL }
\providecommand{\doi}[1]{https://doi.org/#1}

\bibitem{azzopardi2018conceptualizing}
Azzopardi, L., Dubiel, M., Halvey, M., Dalton, J.: Conceptualizing agent-human
  interactions during the conversational search process. In: The second
  international workshop on conversational approaches to information retrieval
  (2018)

\bibitem{bach1979linguistic}
Bach, K., Harnish, R.: Linguistic Communication and Speech Acts. MIT Press
  (1979). \doi{10.2307/2184680}

\bibitem{belkin1980anomalous}
Belkin, N.J.: Anomalous states of knowledge as a basis for information
  retrieval. The Canadian Journal of Information Science  \textbf{5},  133--143
  (1980)

\bibitem{bunt1999dynamic}
Bunt, H.: Dynamic interpretation and dialogue theory. The structure of
  multimodal dialogue  \textbf{2},  139--166 (2000). \doi{10.1075/z.99.10bun}

\bibitem{cohen1960coefficient}
Cohen, J.: A coefficient of agreement for nominal scales. Educational and
  psychological measurement  \textbf{20}(1),  37--46 (1960).
  \doi{10.1177/001316446002000104}

\bibitem{daniels1985using}
Daniels, P.J., Brooks, H.M., Belkin, N.J.: Using problem structures for driving
  human-computer dialogues. In: Recherche d'Informations Assist{\'e}e par
  Ordinateur. pp. 645--660. Centre de hautes études internationales
  d'informatique documentaire (1985). \doi{10.5555/3157680.3157726}

\bibitem{furnas1987vocabulary}
Furnas, G.W., Landauer, T.K., Gomez, L.M., Dumais, S.T.: The vocabulary problem
  in human-system communication. Communications of the ACM  \textbf{30}(11),
  964--971 (1987). \doi{10.1145/32206.32212}

\bibitem{klopfenstein2017rise}
Klopfenstein, L.C., Delpriori, S., Malatini, S., Bogliolo, A.: The rise of
  bots: A survey of conversational interfaces, patterns, and paradigms. In:
  Proceedings of the 2017 conference on designing interactive systems. pp.
  555--565 (2017). \doi{10.1145/3064663.3064672}

\bibitem{mctear2016conversational}
McTear, M.F., Callejas, Z., Griol, D.: The conversational interface, vol.~6.
  Springer (2016). \doi{10.1007/978-3-319-32967-3}

\bibitem{qu2018analyzing}
Qu, C., Yang, L., Croft, W.B., Trippas, J.R., Zhang, Y., Qiu, M.: Analyzing and
  characterizing user intent in information-seeking conversations. In: The 41st
  international acm sigir conference on research \& development in information
  retrieval. pp. 989--992 (2018). \doi{10.48550/arXiv.1804.08759}

\bibitem{radlinski2017theoretical}
Radlinski, F., Craswell, N.: A theoretical framework for conversational search.
  In: Proceedings of the 2017 conference on conference human information
  interaction and retrieval. pp. 117--126 (2017). \doi{10.1145/3020165.3020183}

\bibitem{saracevic1997users}
Saracevic, T., Spink, A., Wu, M.M.: Users and intermediaries in information
  retrieval: What are they talking about? In: User modeling. pp. 43--54.
  Springer (1997). \doi{10.1007/978-3-7091-2670-7_6}

\bibitem{sinclair1975towards}
Sinclair, J.M., Sinclair, J.M., Coulthard, M., et~al.: Towards an analysis of
  discourse: The English used by teachers and pupils. Oxford University Press,
  USA (1975). \doi{10.2307/3585455}

\bibitem{thomas2021theories}
Thomas, P., Czerwinksi, M., McDuff, D., Craswell, N.: Theories of conversation
  for conversational ir. ACM Transactions on Information Systems (TOIS)
  \textbf{39}(4),  1--23 (2021). \doi{10.1145/3439869}

\bibitem{trippas2018informing}
Trippas, J.R., Spina, D., Cavedon, L., Joho, H., Sanderson, M.: Informing the
  design of spoken conversational search: Perspective paper. In: Proceedings of
  the 2018 conference on human information interaction \& retrieval. pp. 32--41
  (2018). \doi{10.1145/3176349.3176387}

\bibitem{trippas2020towards}
Trippas, J.R., Spina, D., Thomas, P., Sanderson, M., Joho, H., Cavedon, L.:
  Towards a model for spoken conversational search. Information Processing \&
  Management  \textbf{57}(2),  102162 (2020). \doi{10.1016/j.ipm.2019.102162}

\bibitem{vakulenko2021large}
Vakulenko, S., Kanoulas, E., De~Rijke, M.: A large-scale analysis of mixed
  initiative in information-seeking dialogues for conversational search. ACM
  Transactions on Information Systems (TOIS)  \textbf{39}(4),  1--32 (2021).
  \doi{10.48550/ARXIV.2104.07096}

\bibitem{vakulenko2020conversational}
Vakulenko, S., Savenkov, V., de~Rijke, M.: Conversational browsing. arXiv
  preprint arXiv:2012.03704  (2020). \doi{10.48550/arXiv.2012.03704}

\bibitem{vtyurina2017exploring}
Vtyurina, A., Savenkov, D., Agichtein, E., Clarke, C.L.: Exploring
  conversational search with humans, assistants, and wizards. In: Proceedings
  of the 2017 chi conference extended abstracts on human factors in computing
  systems. pp. 2187--2193 (2017). \doi{10.1145/3027063.3053175}

\bibitem{white2009exploratory}
White, R.W., Roth, R.A.: Exploratory search: Beyond the query-response
  paradigm. Synthesis lectures on information concepts, retrieval, and services
   \textbf{1}(1),  1--98 (2009). \doi{10.2200/S00174ED1V01Y200901ICR003}

\bibitem{yankelovich2008using}
Yankelovich, N.: Using natural dialogs as the basis for speech interface
  design. In: Human factors and voice interactive systems, pp. 255--290.
  Springer (2008). \doi{10.1007/978-0-387-68439-0_9}

\end{thebibliography}
\end{document}